\documentclass[letterpaper, 10 pt, conference]{ieeeconf}  % Comment this line out if you need a4paper

\usepackage{graphicx}
\usepackage{comment}
\usepackage{amsmath,amssymb} % define this before the line numbering.
\usepackage{color}

\usepackage{algorithm}
\usepackage{algpseudocode}

\usepackage{epsfig}
\usepackage{graphicx}
\usepackage{amsmath,xparse} 
\usepackage{xcolor}

\usepackage{multirow}
\usepackage{multicol}
\usepackage{lipsum}

\usepackage{soul}
\usepackage{xspace}
\makeatletter
\DeclareRobustCommand\onedot{\futurelet\@let@token\@onedot}
\def\@onedot{\ifx\@let@token.\else.\null\fi\xspace}
\def\eg{\emph{e.g}\onedot} 
\def\ie{\emph{i.e}\onedot} 
\def\cf{\emph{cf}\onedot}

\def\etal{\emph{et al}\onedot}

\newcommand{\Tref}[1]{Table~\textcolor{blue}{\ref{#1}}}

\newcommand{\Fref}[1]{Fig.~\textcolor{blue}{\ref{#1}}}
\newcommand{\Sref}[1]{Sec.~\textcolor{blue}{\ref{#1}}}

\IEEEoverridecommandlockouts                              % This command is only needed if 
\title{\LARGE \bf
Stereo Object Matching Network
}

\author{Jaesung Choe$^{1\dagger}$, Kyungdon Joo$^{2\dagger}$, Francois Rameau$^{3}$ and In So Kweon$^{3}$
\thanks{This work (K. Joo) was supported by Institute of Information \& communications Technology Planning \& Evaluation (IITP) grant funded by the Korea government (MSIT) (No.2020-0-01336, Artificial Intelligence Graduate School Program (UNIST)).
This research was also supported by the Shared Sensing for Cooperative Cars Project funded by Bosch~(China) Investment Ltd.
\textit{(Corresponding author: I. S. Kweon.)}
}
\thanks{$^{1}$J. Choe is with the Division of the Future Vehicle, KAIST, Daejeon 34141, Republic of Korea.
{\tt jaesung.choe@kaist.ac.kr}}
\thanks{$^{2}$K. Joo is with the Artificial Intelligence Graduate School and the Department of Computer Science, UNIST, Ulsan 44919, Republic of Korea.
{\tt kdjoo369@gmail.com, kyungdon@unist.ac.kr}}
\thanks{$^{3}$F. Rameau, and I. S. Kweon are with the School of Electrical Engineering, KAIST, Daejeon 34141, Republic of Korea.
{\tt rameau.fr@gmail.com, iskweon77@kaist.ac.kr}}
\thanks{$\dagger$ represents equal contribution.}
}

\newcommand{\FR}[1]{{\textcolor{blue}{FR: #1}}}

\begin{document}

\maketitle
\thispagestyle{empty}
\pagestyle{empty}

%%%%%%%%%%%%%%%%%%%%%%%%%%%%%%%%%%%%%%%%%%%%%%%%%%%%%%%%%%%%%%%%%%%%%%%%%%%%%%%%
\begin{abstract}

This paper presents a stereo object matching method that exploits both 2D contextual information from images as well as 3D object-level information. Unlike existing stereo matching methods that exclusively focus on the pixel-level correspondence between stereo images within a volumetric space~(\emph{i.e.},~cost volume), we exploit this volumetric structure in a different manner. The cost volume explicitly encompasses 3D information along its disparity axis, therefore it is a privileged structure that can encapsulate the 3D contextual information from objects. However, it is not straightforward since the disparity values map the 3D metric space in a non-linear fashion. Thus, we present two novel strategies to handle 3D objectness in the cost volume space: selective sampling~(\emph{RoISelect}) and 2D-3D fusion~(\emph{fusion-by-occupancy}), which allow us to seamlessly incorporate 3D object-level information and achieve accurate depth performance near the object boundary regions. Our depth estimation achieves competitive performance in the KITTI dataset and the Virtual-KITTI 2.0 dataset. 

\end{abstract}

% \keywords{stereo matching, autonomous driving, 3D object detection}

%%%%%%%%%%%%%%%%%%%%%%%%%%%%%%%%%%%%%%%%%%%%%%%%%%%%%%%%%%%%%%%%%%%%%%%%%%%%%%%%

% =======================================
% Introduction
% =======================================
\vspace{+3mm}
\section{Introduction}
%
% Stereo matching is a fundamental task in computer vision which consists of reconstructing the 3D environment.
%
%Stereo matching is a fundamental task in computer vision which consists of \FR{consists in --> https://blog.harwardcommunications.com/2014/06/11/the-difference-between-consist-of-and-consist-in/} 
Stereo matching is a fundamental task in computer vision which consists in
reconstructing the 3D environment captured from a pair of cameras.
Specifically, stereo matching refers to the process of computing the dense pixel correspondence between a rectified image pair~\cite{geometry}. 
This correspondence is expressed as the horizontal pixel distance from one pixel on the left image to its corresponding position in the right image. 
%
% This distance is known as \emph{disparity} and this disparity map \FR{what disparity map, it has not been introduce. ". This disparity can be estimated for each pixel in the image, leading to a disparity map which can be utilized for various computer vision tasks, such as ..."} can be utilized in various computer vision tasks, such as semantic segmentation~\cite{segstereo,semantic-costV}, stereo-LiDAR fusion~\cite{stereolidar-jaesungchoe}, and 3D object detection tasks~\cite{pseudo_lidar,monodepthdetection,pseudo_lidar++}.
%
This distance is known as \emph{disparity}. This disparity can be estimated for each pixel in the image, leading to a disparity map which can be utilized for various computer vision tasks, such as semantic segmentation~\cite{segstereo,semantic-costV}, stereo-LiDAR fusion~\cite{stereolidar-jaesungchoe}, and 3D object detection tasks~\cite{pseudo_lidar,monodepthdetection,pseudo_lidar++}.

Recently, stereo-vision has benefited from the advances of Convolutional Neural Networks (CNNs)~\cite{alexnet,vgg,resnet}. 
%
% The pioneering deep learning-based patch matching approach~\cite{mccnn} largely outperforms the traditional hand-crafted techniques~\cite{semi_global_matching}. 
% \FR{"Pioneering deep learning-based patch matching approaches, such as~\cite{mccnn}, largely outperform traditional hand-crafted techniques~\cite{semi_global_matching}"} 
Pioneering deep learning-based patch matching approaches, such as~\cite{mccnn,dispnet}, largely outperform traditional hand-crafted techniques~\cite{semi_global_matching}.
Follow-up deep learning-based solutions, inspired by conventional stereo matching techniques~\cite{stereo-taxonomy}, explicitly incorporate geometric information in the matching process~\cite{gcnet}, which produces
%  \FR{"leading to" instead of "which produces"}
significant accuracy improvements.
For example, Kendall~\etal~\cite{gcnet} introduce a differentiable cost volume to compute matching costs between the left and right features from the stereo images. 
With its larger receptive field and larger capacity, this cost volume approach has demonstrated highly accurate results. This seminal work has inspired various methods that rely on cost volume~\cite{gcnet,psmnet,ganet,cspn}.
%
% This seminal work has inspired various methods that rely on cost volume~\cite{gcnet,psmnet,ganet,cspn}.

\begin{figure}[!t]
\centering
\includegraphics[width=0.65\linewidth]{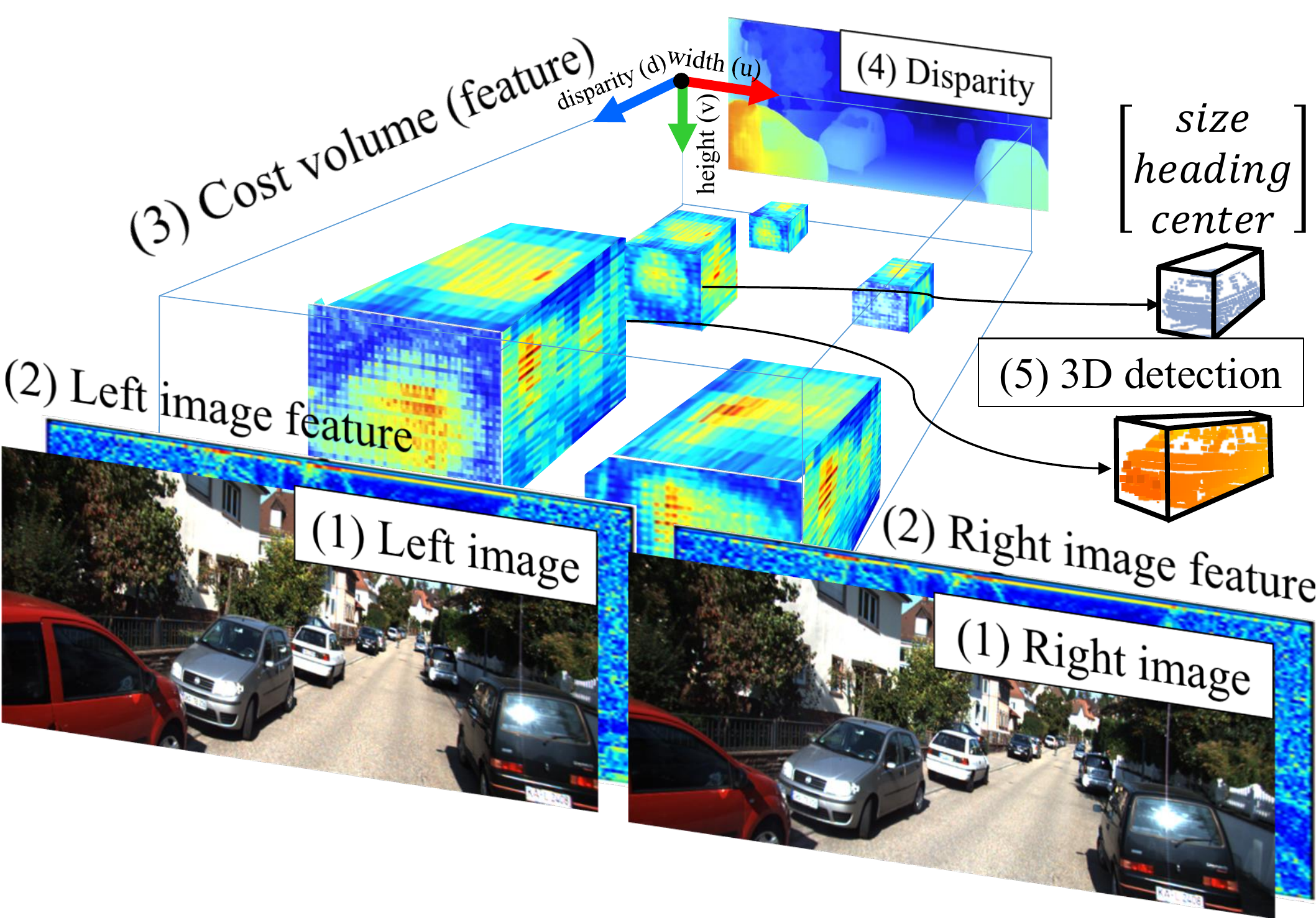}
\vspace{-2mm}
\caption[Caption for LOF]{\textbf{Overview.} Given (1)~stereo images,
we extract the (2)~left/right feature maps~($\mathcal{F}_{L}, \mathcal{F}_{R}$)
to construct the (3)~cost volume $\mathcal{V}$.
Our cost volume is aggregated to~(4) estimate the disparity map 
and to (5)~detect 3D object.
}
\label{fig:main}
\vspace{-5mm}
\end{figure}

Despite the improvements arising from CNNs, multiple challenges still remain. Particularly, textureless or repetitive structures in the image lead to inaccurate disparity estimation. 
Moreover, objects' boundaries tend to be ambiguous. 
To overcome these issues, several methods apply multi-task losses using additional information, such as image segmentation~\cite{semantic-costV}. 
These approaches rely on 2D contextual information extracted from the image. These 2D cues improve the matching quality but are inherently ambiguous since contiguous pixels in the image space are not necessarily neighbors in 3D.
In contrast, considering an object prior in 3D can reduce the object-level spatial ambiguity effectively. Motivated by this observation, we investigate how such a geometric property can be integrated into a stereo matching network.

In this paper, we propose a stereo matching network that utilizes 2D information as well as 3D contextual information of objects. We named this method \emph{stereo object matching network} (see \Fref{fig:main}).
Our method builds on a cost volume-based approach~\cite{psmnet}, which estimates a depth map using 2D contextual information and includes an implicit 3D space structure via the cost volume.
Concretely, we perform a 3D object detection task, as an auxiliary task, using the cost volume to guide the network to learn meaningful structural information about the objects in the scene. 
This strategy reduces the uncertainty of the disparity boundary near the objects (\eg,~vehicle).
Therefore, the cost volume in our network considers both conventional disparity loss in a pixel-wise manner and detection loss from the 3D RoIs which are extracted from the cost volume.
Each element in the multi-task losses has geometric-awareness and makes it possible to achieve state-of-the-art performance for depth estimation in the KITTI dataset~\cite{kitti} and Virtual-KITTI dataset~\cite{virtual-kitti}.

% =======================================
% Related Work
% =======================================
\section{Related work}

Stereo matching has been deeply investigated in the past decades.
Recently, deep learning techniques paved the way for more effective and robust matching. 
The work by Zbontar and LeCun~\cite{mccnn} is the first technique to employ deep-patch-based stereo matching and 
Mayer~\etal~\cite{dispnet} introduce an end-to-end architecture to regress the disparity.
More effective and elegant approaches inspired by traditional matching techniques have been developed. It is, for instance, the case of Kendall~\etal~\cite{gcnet} who propose to integrate a differentiable cost volume into a deep neural architecture. 
Relying on this concept, several works enhance the deep architecture by widening the receptive field using a pyramid network~\cite{psmnet},
exploiting the semi-global cost aggregations~\cite{ganet},
and propagating the spatial pixel-affinity modules~\cite{cspn}.
However, the objects' boundary ambiguity is a challenging issue 
and such ambiguity results in a low-quality depth prediction on these areas.
Recent works~\cite{segstereo,semantic-costV,miclea2019real,sun2020real} 
integrate semantic segmentation into the disparity estimation to handle this ambiguity issue, but the resulting improvements remain relatively limited due to the inherent ambiguity of this semantic two-dimensional information.

Meanwhile, camera-based 3D object detection increases the performance by exploiting high-quality depth from deep learning-based approaches~\cite{monodepth,dispnet}.
Xu and Chen~\cite{monodepthdetection} propose a multi-level fusion technique and Wang~\etal~\cite{pseudo_lidar} convert the depth into pseudo lidar data for vehicle localization.
These approaches demonstrate that the quality of the depth affects the performance of the detection. However, they only utilize depth information as an additional modality in a naive manner (\eg,~concatenate with input).

In this study, we directly handle the 3D contextual information by exploiting 3D object detection as an auxiliary task.
Our method utilizes 3D objects' structural information (\eg,~size and heading) so that our cost volume is trained to express the detection-based 3D contextual information.
By our cost volume-based detection network, 
the proposed approach largely increases the quality of estimated depth.

%%%%%%%%% Fig-arch (2-column) %%%%%%%%% 
% \begin{figure*}[t]
% \centering
% \includegraphics[width=0.8\linewidth]{figure/archFig/fig_architecture_05.pdf}
% \vspace{-1mm}
% \caption{\textbf{Illustration of stereo object matching network.}
% Our network consists of three parts: (1) cost volume prediction, (2) 3D object detection, and (3) disparity regression. 
% The refined cost volume $\mathcal{V}$ is utilized to regress disparity $\widetilde{\mathcal{D}}$ and to detect the 3D objects by multi-task losses that simultaneously affect cost volume.
% We visualize the attention of features ($\mathcal{F}_{L}$, $\mathcal{F}_{R}$, $\mathcal{V}$) and cost aggregation $\mathcal{A}$.
% We also illustrate the location of the 3D RoI ($\{ \mathbf{p}_{min}, \mathbf{p}_{max} \}$) by the 3D-RPN as the red 3D box in $\mathcal{V}$.
% }
% \label{fig:architecture}
% \vspace{-1\baselineskip}
% \end{figure*}

%%%%%%%%% Fig-arch (1-column) %%%%%%%%% 
\begin{figure}[t]
\centering
\includegraphics[width=1.0\linewidth]{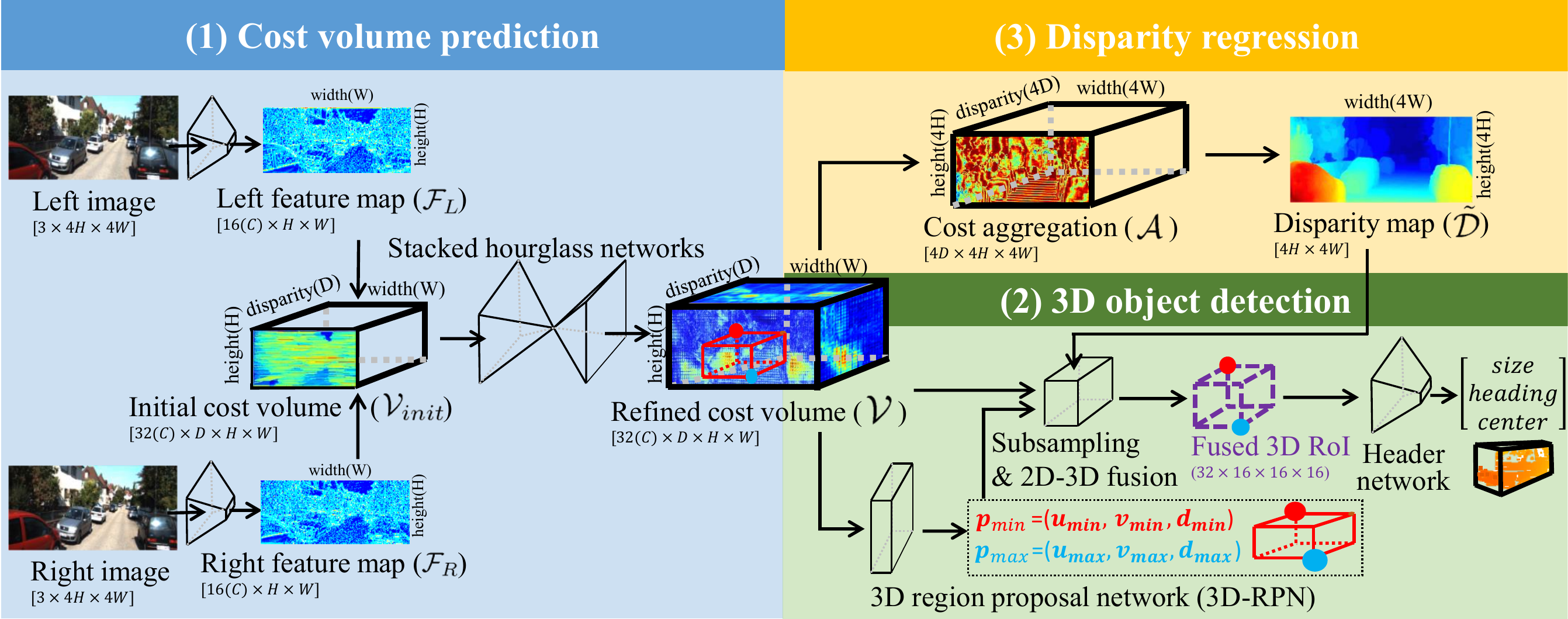}
% \vspace{-1mm}
\caption{\textbf{Illustration of stereo object matching network.}
Our network consists of three parts: (1) cost volume prediction, (2) 3D object detection, and (3) disparity regression. 
The refined cost volume $\mathcal{V}$ is utilized to regress disparity $\widetilde{\mathcal{D}}$ and to detect the 3D objects by multi-task losses that simultaneously affect cost volume.
We visualize the attention of features ($\mathcal{F}_{L}$, $\mathcal{F}_{R}$, $\mathcal{V}$) and cost aggregation $\mathcal{A}$.
We also illustrate the location of the 3D RoI ($\{ \mathbf{p}_{min}, \mathbf{p}_{max} \}$) by the 3D-RPN as the red 3D box in $\mathcal{V}$.
}
\label{fig:architecture}
% \vspace{-1\baselineskip}
\end{figure}

% =======================================
% Stereo Object Matching
% =======================================
\section{Stereo Object Matching}

In this section, we propose a stereo object matching network, which consists of three steps: cost volume prediction, disparity regression, and 3D object detection. Each module of the pipeline is introduced individually, and the overall architecture is depicted in~\Fref{fig:architecture}.

\subsection{Overview}

Given a rectified stereo pair, we propose a disparity estimation method, called \emph{stereo object matching network}, which considers 2D contextual information as well as 3D object-level information. We first extract image features from stereo images and then construct a cost volume~(\Sref{subsection : cost volume}). 
The conventional cost volume is used to predict the disparity map out of 2D image information only.
However, since the image features are concatenated and stacked for each disparity hypothesis, this cost volume follows the 3D voxel structure with feature vectors, \ie, channel~(C)$\times$disparity~(D)$\times$height~(H)$\times$width~(W). 

In this work, we attempt to fully exploit this 3D voxel structure not only to compute \FR{the} disparity but also to embed 3D object-level information via an auxiliary 3D detection task (we discuss the details of how we embed 3D object-level contextual information in \Sref{section : Detection-based 3D contextual embedding}). 
It allows us to utilize both 2D contextual and 3D object-level information, which produces accurate disparity estimation~(\Sref{subsection : disparity network}).

\subsection{Cost volume}
\label{subsection : cost volume}

From rectified stereo images, we first extract the stereo image feature maps, \ie, left feature map~$\mathcal{F}_L$ and right feature map~$\mathcal{F}_R$ whose size is 16$\times$H$\times$W using the feature extraction layer in PSMNet~\cite{psmnet}.
A strong similarity between two feature vectors from each feature map indicates a high matching probability, 
meaning that two pixels corresponding to the two feature vectors are the projection of the same 3D point in the scene. 

The initial cost volume~$\mathcal{V}_{init}$ is built as follows~\cite{gcnet,psmnet}: 
(1)~shifting right image features horizontally with fixed left image features (each shift corresponding to a candidate disparity value),
(2)~stacking the shifted right and fixed left features across each disparity level,
and (3)~concatenating the left and right features along the channel axis.
This initial cost volume is then refined via stacked hourglass networks~\cite{psmnet,associate_embedding,hourglass}. These stacked hourglass networks are important to regularize the initial cost volume and to increase the receptive field of the network. 
The constructed cost volume~$\mathcal{V}$ forms a 4D volume\footnote{Although the physical dimension of $\mathcal{V}$ is ``C$\times$D$\times$H$\times$W'', we can consider $\mathcal{V}$ as a 3D volume with its size ``D$\times$H$\times$W'', where the C-dimensional feature vector is stored at each voxel's location $(u,v,d)$. Thus, we interchangeably describe $\mathcal{V}$ as 3D volume or 4D volume in some cases.}, {which describes the 3D space along the disparity axis, and where high responses are clustered object-wise (see the visualization of $\mathcal{V}$ in \Fref{fig:architecture}).}
This cost volume is the central element of our system since it is explicitly used for both disparity estimation and 3D object detection, especially as an intermediate space to embed 3D object-level information.

\subsection{Disparity network} \label{subsection : disparity network}

Given the learned cost volume $\mathcal{V}$ that embeds both 2D and 3D contextual information, the disparity network aims to regress the disparity values at a subpixel level.
Following~\cite{gcnet}, we perform a cost aggregation process by aggregating the cost volume along the disparity dimension as well as its spatial dimensions.
This cost aggregation reduces the dimensionality of $\mathcal{V}$ into a 3D structure $\mathcal{A}$.
From $\mathcal{A}$, we can estimate the disparity value $\tilde{d}_{u,v}$ at pixel $(u,v)$ as: \vspace{-2mm}
\begin{equation}
    \tilde{d}_{u,v} = \sum_{d=0}^{d_{max}} d \cdot \sigma ({\mathbf{a}}_{u,v}^d),
    \label{equation:intra-pixel}
    \vspace{-2mm}
\end{equation} 
where $d_{max}$ indicates the maximum disparity, $\sigma(\cdot)$ denotes softmax operation,
and ${\mathbf{a}}_{u,v}^d$ is the response for disparity value $d$ 
of the cost aggregation vector ${\mathbf{a}}_{u,v}$ $\in \mathbb{R}^{d_{max}}$ at $(u,v)$.

We compute the disparity loss~$\mathcal{L}_{disp}$ from the estimated disparity map~$\tilde{\mathcal{D}}$ and the ground truth disparity map~$\mathcal{D}$ as: \vspace{-1mm}
\begin{equation}
    \mathcal{L}_{disp} = \frac{1}{N}\sum_{u}\sum_{v} smooth_{L_{1}}({d}_{u,v} - \tilde{d}_{u,v}),
    \vspace{-2mm}
    \label{equation:disp-loss}
\end{equation}
where $N$ is the number of the pixels in $\mathcal{D}$, $\tilde{d}_{u,v}$ is the value of the predicted disparity map $\widetilde{\mathcal{D}}$ at the pixel $(u, v)$,
and $smooth_{L_{1}}(\cdot)$ is the smooth L1 loss function used to compute the loss~\cite{gcnet,psmnet}.

% =======================================
% Detection-based 3D contextual embedding
% =======================================
\section{Detection-based 3D contextual embedding} 
\label{section : Detection-based 3D contextual embedding}

In this section, we propose a cost volume-based 3D object detection, as an auxiliary task to embed 3D object information. 
We exploit an anchor-based detection framework~\cite{faster_rcnn} composed of two stages -- a 3D region proposal network (3D-RPN) and a header network.
While previous techniques have been developed for 2D input~\cite{stereo_rcnn,mutlibox3d}, our 3D object detection uses cost volume as input. 
Directly using the cost volume for 3D object detection is meaningful since it encapsulates both 2D and 3D information; however, it remains a complex task. 
The challenges of using such a structure to perform 3D object detection, as well as the proposed solutions (selective subpixel sampling and 2D-3D fusion), are discussed below.

\begin{figure}[t]
\centering
\includegraphics[width=1.00\linewidth]{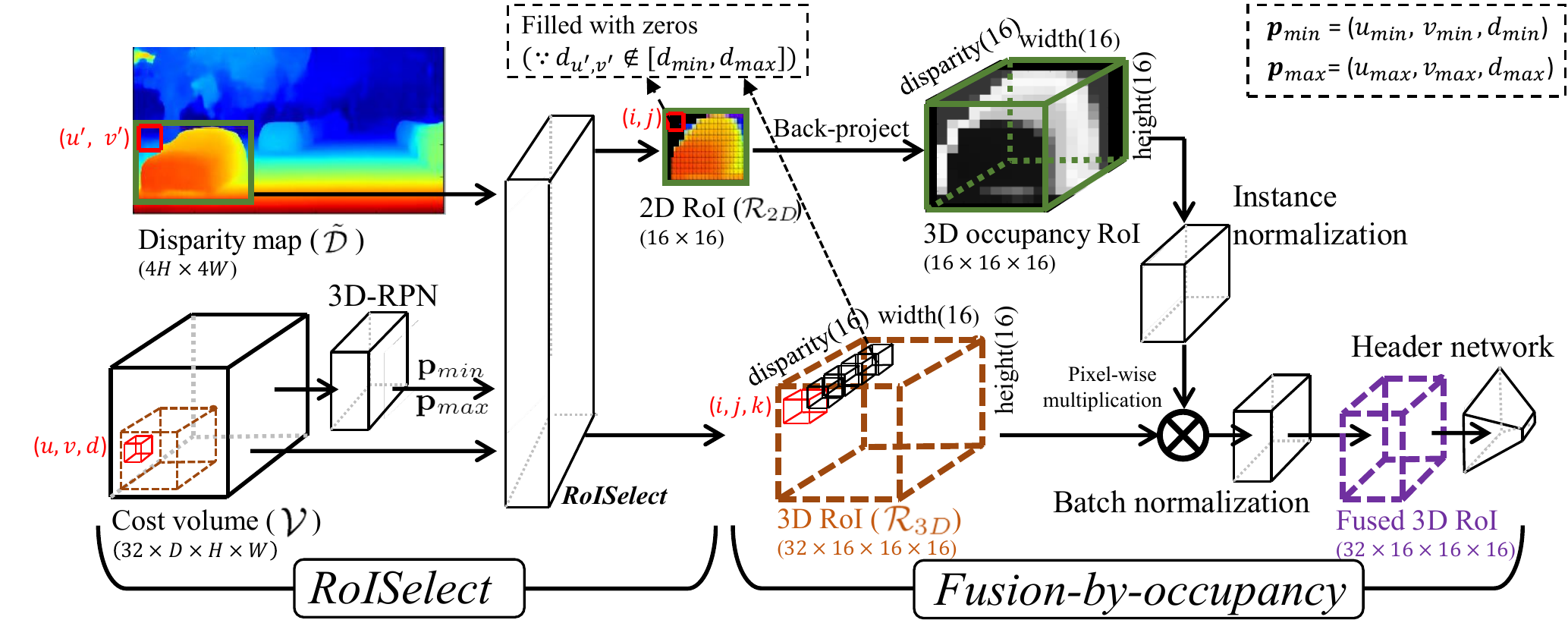}
\vspace{-3mm}
\caption{\textbf{Illustration of the RoISelect and the fusion-by-occupancy.}
    Given the cost volume $\mathcal{V}$, the estimated disparity $\widetilde{\mathcal{D}}$, and the extracted 3D RoI $\{ \mathbf{p}_{min}, \mathbf{p}_{max} \}$ by 3D-RPN, 
    RoISelect estimates the subsampled 2D and 3D RoIs: 2D RoI $\mathcal{R}_{2D}$ (in the estimated disparity map) and 3D RoI $\mathcal{R}_{3D}$ (in the cost volume), 
    where we deal with unreliable 2D pixel/3D voxel (\eg, background or occlusion region) by filling the corresponding location as zero.
    For the estimated 2D RoI $\mathcal{R}_{2D}$, we apply the fusion-by-occupancy that estimates 3D occupancy RoI by back-projecting $\mathcal{R}_{2D}$ into 3D RoI aligned with $\mathcal{R}_{3D}$.
    After RoISelect and fusion-by-occupancy, we fuse the 3D RoI and the 3D occupancy RoI using the normalization layers~\cite{batch_norm,inst-norm}. The fused 3D RoI is fed into the header network.
}
\label{fig:fusion}
\vspace{-1\baselineskip}
\end{figure}

% \subsection{Metric-driven 3D structure embedding}
\subsection{Metric-driven cost volume}
\label{subsection:metric-driven 3d structure embedding}

By virtue of its structure (\ie,~3D volumetric space along the disparity axis), the cost volume naturally encodes the 3D space.
However, utilizing this volume as a direct input for 3D object detection is not straightforward for two major reasons.
1)~The disparity is represented in the discrete pixel domain, which is not linearly correlated with metric depth.
For example, a disparity error of one pixel for nearby objects represents a metric error of a few centimeters, while the same discrepancy for faraway objects can result in a significant error (up to a few meters depending on the camera resolution and baseline's length). In other words, an extracted anchor corresponding to a distant object may include only a small region in the cost volume while a near object of physically similar size will appear significantly larger.
2)~A~sampled cost volume usually includes uncertain matching costs due to occlusions or the limited overlapping area between both views.
{Valid matching costs (\ie,~high confidence) only exist where the camera can see. That is, we cannot compute valid matching costs beyond the visible surface.}
Therefore, wrongly estimated matching costs on an occluded region might confuse the network to incorrectly localize the objects, which affects the cost volume during training.

To resolve these issues on the cost volume, we propose two strategies (see \Fref{fig:fusion}):
1)~A selective subsampling method, called \emph{RoISelect}, with the aid of subpixel interpolation, 
and 2)~2D-3D fusion using occupancy reconstruction, called \emph{fusion-by-occupancy} to inject the continuous {depth} values into the discrete 3D RoI from the cost volume.

\begin{figure}[!t]
    \vspace{-2mm}
    \centering
    \includegraphics[width=0.99\linewidth]{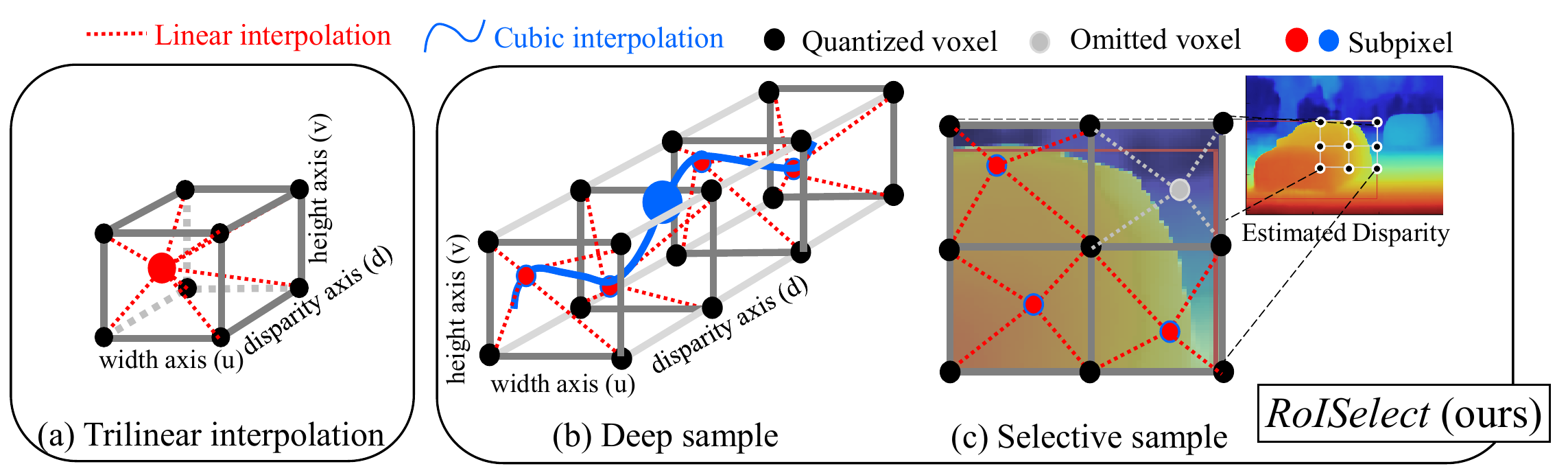}\\
    \vspace{-3mm}
    \caption{\textbf{Illustration of the RoI sampling strategies.}
    Compared to (a) trilinear interpolation, our \emph{RoISelect} consists of (b) Deep sample module and (c) Selective sample module,
    This is to sample the valid voxels in cost volume by using disparity.}
    \label{fig:sampling}
\vspace{-1\baselineskip}
\end{figure}

\vspace{1mm} \noindent \textbf{Selective subpixel sampling (RoISelect).} \
In the anchor-based detection framework,
since 3D-RPN extracts 3D RoI using pre-defined anchors, the sampling process is essentially required to normalize the RoI to a common size.
In particular, due to the above mentioned potential issues of the cost volume, extracted 3D RoI in the cost volume space has a highly varying size 
and uncertain matching costs depending on the location and occlusion.
Thus, we propose a selective subsampling method, called \emph{RoISelect}, which selects high confidence voxels using the estimated disparity map~(\cf,~\Sref{subsection : disparity network}) and interpolates (samples) them while considering the continuous metric range as depicted in Figs.~\ref{fig:sampling}--(\textcolor{blue}{b},~\textcolor{blue}{c}).
Our subsampling method functions in a way similar to a linear interpolation
(\eg,~\emph{RoIAlign}~\cite{mask_rcnn}).
However, we broaden the receptive field of sampling by applying the cubic interpolation only to the disparity axis, 
called a deep sample module as in~\Fref{fig:sampling}--(\textcolor{blue}{b}).
In addition, we filter the low confidence voxels that belong to the background disparity information, as shown in~\Fref{fig:sampling}--(\textcolor{blue}{c}). For instance in~\Fref{fig:sampling}--(\textcolor{blue}{c}), the top-right voxel is determined to be omitted because disparity map implies that the high confidence voxels are located at the background area (\ie, out of the 3D RoI).
In this case, the omitted voxels do not propagate the gradient to the cost volume (similar to ReLU~\cite{deep-learning-book}) so that we can regularize the cost volume along the disparity-axis\footnote{Details of \emph{RoISelect} is described in the supplementary material.}.
Thanks to our {\emph{RoISelect}}, we can selectively subsample the 3D RoIs that contain the high confidence voxels from the cost volume.

%%%%% DO NOT ERASE IT
{
}

\vspace{1mm} \noindent \textbf{2D-3D fusion~(fusion-by-occupancy).} \
Fusion with metric measurement~(\eg, depth) is widely used in camera-based 3D object detection methods~\cite{monodepthdetection,seg2reg,roi-10d}.  
These approaches blend 2D contextual features with depth using the multi-level injection~\cite{monodepthdetection} or normalization layers~\cite{seg2reg} to accurately localize the target objects. 
For our cost volume, it is necessary to match the dimensions of both 3D RoI~$\mathcal{R}_{3D}$ (from the cost volume) and 2D RoI~$\mathcal{R}_{2D}$ (from the estimated disparity map) for aligned fusion.
There are several ways to encapsulate the estimated depth into the 3D structure.
We propose fusion-by-occupancy to properly fuse the depth into the 3D feature.

As illustrated in~\Fref{fig:fusion}, 
we back-project the 2D RoI~$\mathcal{R}_{2D}$ using its depth measurements into the 3D space as a binary representation of the object surface, \ie, 3D occupancy RoI.
In detail, we assign 1 for the back-projected voxels in the 3D occupancy RoI and fill other voxels with 0.
By doing so, we only propagate the backward gradient through the valid voxels in the 3D RoI. This is to regularize the proper voxels in the cost volume along the disparity axis, as \emph{RoISelect} does.
By reconstructing the 3D surface through our fusion-by-occupancy, we preserve the valid feature voxels within the 3D RoI.
After the reconstruction, we fuse the 3D RoI~$\mathcal{R}_{3D}$ with the 3D occupancy RoI using the normalization layers.
With fusion-by-occupancy, we can align the dimensions of the 2D depth and the 3D feature. Accordingly, we can increase the accuracy of the 3D object detection and the quality of the depth estimation. We validate the fusion-by-occupancy in an ablation study~(see~\Tref{table:ablation-fusion}).

\subsection{Cost volume-based 3D object detection}
\label{subsection : Cost volume-based 3D object detection}
We adopt the anchor-based detection (\ie,~two-stage detection) as an auxiliary task to guide the 3D contextual information to the cost volume through the pre-defined anchors.
In addition, the subsequent header network provides more detailed contextual information, such as the heading and the size of vehicles.

\vspace{1mm} \noindent \textbf{3D region proposal network.} \
To seamlessly connect the pre-defined anchors with our cost volume~$\mathcal{V}$, we exploit a 3D region proposal network (3D-RPN) defined in the cost volume coordinate, \ie, $(u, v, d)$ coordinate. 
Given the cost volume~$\mathcal{V}$, the 3D-RPN is trained to (1)~predict the potential location of the objects with respect to the pre-defined anchors, and (2)~classify the foreground anchors.
The 3D-RPN loss $\mathcal{L}_{rpn}$ consists of the two typical losses from the 3D-RPN~\cite{faster_rcnn} as: \vspace{-1mm}
\begin{equation}
    \mathcal{L}_{rpn} = \mathcal{L}_{class} + \mathcal{L}_{anc},
    \vspace{-1mm}
    \label{equation:rpn-loss}
\end{equation}
where $\mathcal{L}_{class}$ is the binary cross-entropy loss for the foreground classification and $\mathcal{L}_{anc}$ is the anchor localization loss.
The 3D-RPN estimates the location of the 3D RoIs $\{ \mathbf{p}_{min}, \mathbf{p}_{max} \}$ in the cost volume space, as shown in~\Fref{fig:architecture}.

\vspace{1mm} \noindent \textbf{Header network.} \
Given $\mathcal{R}_{3D}$ with a fixed size (by RoISelect and fusion-by-occupancy), we design the header network to predict the center location $(u_{obj}, v_{obj}, d_{obj})$, size $(w_{obj}, h_{obj}, l_{obj})$, and heading $(yaw_{obj})$ of the objects.
The header network is trained to regress the objects' parameters with respect to the location of the 3D RoIs. 

We follow up the overall architecture proposed by Mousaivan~\etal~\cite{mutlibox3d} and Xu and Chen~\cite{monodepthdetection}, except for the heading prediction. 
Recent camera-based 3D object detection methods~\cite{mutlibox3d,monodepthdetection} adopt the alpha observation (\ie, local orientation)~\cite{kitti,mutlibox3d} as the heading of the vehicles.
%
% The reason \FR{what reason? I do not understand this sentence} is that their feature representation is limited to the 2D contextual attentions~(C$\times$H$\times$W) such that the same headed vehicles can be differently visualized depending on their distance to the camera~\cite{mutlibox3d}, which shows a tendency similar to the alpha observation. 
Because their feature representations is limited to the 2D contextual attentions~(C$\times$H$\times$W), the same headed vehicles can be differently visualized depending on their distance to the camera~\cite{mutlibox3d}, which shows a tendency similar to the alpha observation.
However, in our case, the cost volume inherently expresses the 3D structures along the disparity axis.
Thus, we estimate the original heading of the vehicle (\ie,~yaw rotation) from the 3D RoIs, instead of the alpha observation. We cover the details of the architecture of the header network in the supplementary material.

Finally, our cost volume is additionally trained by the loss from the detection networks.
The loss function of our stereo object matching network is:\vspace{-1mm}
\begin{equation} 
    \mathcal{L}_{total} = \mathcal{L}_{disp} + \mathcal{L}_{rpn} + 2 \cdot \mathcal{L}_{header},
    \vspace{-1mm}
    \label{equation:total-loss}
\end{equation}
where $\mathcal{L}_{header}$ is the loss from the header network, in which we apply (1)~the L1-loss for the regression of the size, heading, and the center location of the objects, and (2)~the binary cross-entropy loss for the confidence prediction. 
We further describe our loss design in the supplementary material.

%------------------------------------------------------------------------

% =======================================
% Experimental Result
% =======================================

\section{Experimental Result}
We describe the training process using the KITTI dataset~\cite{kitti} and evaluate the accuracy of the depth~(disparity) and detection against a large panel of existing techniques in the KITTI dataset~\cite{kitti} and Virtual-KITTI 2.0 dataset~\cite{virtual-kitti}.
Note that we carefully filtered out the training data to ensure a fair comparison with the depth estimation networks~\cite{eigen_mono_depth,monodepth,psmnet,pseudo_lidar}.
In addition, we conduct an ablation study to validate the different modules of our method.

\subsection{Implementation}
\label{subsection:implementation}
Our training scheme consists of three steps.
First, we pre-train the cost volume prediction part and the disparity regression network with Scene Flow dataset~\cite{dispnet} by setting the learning rate to 0.001 during the 15 epochs.
Second, we continue to train the cost volume prediction network and the disparity regression network 
with the KITTI-split by Eigen~\etal~\cite{eigen_mono_depth} by setting the learning rate to 0.0005 for 30 epochs. More details are included in the supplementary material.

\begin{comment}
To guarantee a fair comparison, we exclude the part of the training data that are included in the KITTI object validation dataset~\cite{kitti-object} and the KITTI stereo train-val dataset~\cite{kitti-stereo}.
Moreover, we use identical/similar hyper-parameters that recent methods used~\cite{psmnet,pseudo_lidar,DSGN}: the dimensions of cost volume ($C({=}48) \times D({=}48) \times H \times W$). For training, we apply random crop augmentation so the resulting size of input images is $256 \times 512$. For inference, we use the original size of the input images as the recent methods do~\cite{psmnet,pseudo_lidar,DSGN}.
Last, we finetune the whole network~(including the 3D-RPN and header network) with the KITTI object training dataset~(learning rate is set as 0.0001 for 20 epochs).
Since the object's label information, such as the heading and the size, is only provided within the KITTI object benchmark, the last training phase is the key to our training session.

In Virtual-KITTI 2.0 dataset~\cite{virtual-kitti}, we follow the evaluation scheme in Cabon~\etal~\cite{virtual-kitti}; we do not finetune the network and exploit the pre-trained weights on the KITTI dataset. Typically, we use part of the data (15-deg-left) for depth evaluation, totally 2,303 images, to test the generalization performance of the network.
\end{comment}

%%%%%%%%%%%%% KITTI depth (2-column) %%%%%%%%%%%%%%%%%%%%%%%%%%%%%%%%%%%%
{
}
%%%%%%%%%%%%% KITTI depth (1-column) %%%%%%%%%%%%%%%%%%%%%%%%%%%%%
\begin{table}[t!]
\centering
\caption{\textbf{Quantitative results for depth estimation in the KITTI split by Eigen~\etal~\cite{eigen_mono_depth}.} 
Stereo matching techniques usually provide disparity estimation (not depth), 
while monocular-based studies are mainly evaluated for depth estimation.
Thus, we provide comparisons on both monocular and stereo approaches for a complete evaluation on a unified unit -- depth metric. 
* indicates the reproduced results. 
M and S mean mono camera and stereo camera, respectively.
}
\vspace{-2mm}
\resizebox{0.935\linewidth}{!}{
\begin{tabular}{|c|c|c|c|c|}
\hline
\multirow{3}{*}{Method} & \multicolumn{1}{c|}{\multirow{3}{*}{Modality}} & \multicolumn{3}{c|}{Depth Evaluation} \\ 
& \multicolumn{1}{c|}{} & \multicolumn{3}{c|}{(Lower the better)}\\ \cline{3-5} 
& \multicolumn{1}{c|}{} & Abs Rel & Sq Rel & RMSE\\ \hline
Eigen~\cite{eigen_mono_depth} & M & 0.203 & 1.548 & 6.307 \\ \hline

DORN~\cite{dorn} & M & 0.072 & 0.307 & 2.727 \\ \hline\hline

PSMnet~\cite{psmnet}* & S & 0.053 & 0.234 & 2.847 \\ \hline

PseudoLiDAR~\cite{pseudo_lidar} & S & 0.052 & 0.281 & 3.027 \\ \hline

DSGN~\cite{DSGN} & S & 0.064 & 0.184 & 2.942 \\ \hline

% PseudoLiDAR\texttt{++}~\cite{pseudo_lidar++} & S & 0.059 & 0.165 & 1.992 \\ \hline\hline

Ours & S & \textbf{0.027} & \textbf{0.111} & \textbf{1.842} \\ \hline
\end{tabular}
}
\label{table:kitti-eigen-metric}
% \vspace{-2mm}
\end{table}

{
}
%%%%%%%%%%%%% Virtual-KITTI depth (1-column) %%%%%%%%%%%%%%%%%%%%%%%%%%%%%
\begin{table}[t!]
\centering
\caption{\textbf{Quantitative results for depth estimation in Virtual-KITTI dataset.} 
* means the reproduced results. M and S indicate mono camera and stereo camera, respectively.
}
\vspace{-2mm}
\resizebox{0.935\linewidth}{!}{
\begin{tabular}{|c|c|c|c|c|}
\hline

\multirow{3}{*}{Method} & \multirow{3}{*}{Modality} & \multicolumn{3}{c|}{Depth Evaluation} \\ 

& \multicolumn{1}{c|}{} & \multicolumn{3}{c|}{(Lower the better)} \\ \cline{3-5}

& \multicolumn{1}{c|}{} & Abs Rel & Sq Rel & RMSE \\ \hline

PSMnet~\cite{psmnet}* & S & 0.050 & 0.517 & 4.909 \\ \hline

PseudoLiDAR~\cite{pseudo_lidar}  & S & 0.080 & 1.044 & 5.137 \\ \hline

% PseudoLiDAR\texttt{++}~\cite{pseudo_lidar++}  & S & 0.057 & \textbf{0.488} & 4.650 \\ \hline

DSGN~\cite{DSGN}  & S & 0.059 & \textbf{0.503} & 4.748 \\ \hline\hline

Ours & S & \textbf{0.044} & 0.557 & \textbf{4.619} \\ \hline

\end{tabular}
}
\vspace{-1\baselineskip}
\label{table:virtual-kitti-metric}
\end{table}

% %%%%%%%%%%%%%%%%%%%% KITTI-detection (2-column) %%%%%%%%%%%%%%%%%%%%%%%%%%%%%
{
}
%%%%%%%%%%%%%%%%%%%% KITTI-detection (1-column) %%%%%%%%%%%%%%%%%%%%%%%%%%%%%
\begin{table}[t!]
\centering
\caption{\textbf{Quantitative results for 2D and 3D detection in the KITTI object validation benchmark.} 
Though our method focuses on depth estimation, we provide the accuracy of our detection results (\ie,~3D-RPN and header). 
Our method shows relatively lower accuracy compared to the recent methods. 
Our reasoning is that the cost volume encompasses the 3D space in disparity axis, 
which is the weak point in detection 3D object, 
while the similar approach~\cite{DSGN} transforms the cost volume into the metric space volume and achieves the highest accuracy in 3D detection task. 
M and S mean mono camera and stereo cameras, respectively.}
\resizebox{1.0\linewidth}{!}{
\begin{tabular}{|c|c|c|c|c|}
\hline

% \multirow{3}{*}{Method} & \multirow{3}{*}{Modality} & 2D Detection & \multicolumn{2}{c|}{3D Detection} \\ \cline{3-5} 
% & & $\text{AP}_{\text{2D}}$ & $\text{AP}_{\text{BEV}}$ & $\text{AP}_{\text{3D}}$ \\ 

\multirow{2}{*}{Method} & \multirow{2}{*}{Modality} & $\text{AP}_{\text{2D}}$ & $\text{AP}_{\text{BEV}}$ & $\text{AP}_{\text{3D}}$ \\ % \cline{3-5} 

& & (Moderate) & (Moderate) & (Moderate) \\ \hline
Multi-Fusion~\cite{monodepthdetection} & M & -  & 13.63 & 5.69 \\ \hline

M3D-RPN~\cite{mono_rpn}                & M & 83.67 & 21.18 & 17.06 \\ \hline\hline
3DOP~\cite{stereo3d}                   & S & 88.07 & 9.49 & 5.07 \\ \hline
TLnet~\cite{triangulation_detection}   & S & - & 21.88 & 14.26 \\ \hline
StereoRCNN~\cite{stereo_rcnn}          & S & \textbf{88.27} & 48.30 & 36.69 \\ \hline
PseudoLiDAR~\cite{pseudo_lidar}        & S & - & 56.80 & 45.30 \\ \hline
DSGN~\cite{DSGN}                       & S & 83.59 & \textbf{63.91} & \textbf{54.27} \\ \hline\hline
Ours                                   & S & 76.71 & 30.92 & 19.88 \\ \hline
\end{tabular}
}
\label{table:kitti-object-metric}
\end{table}

%%%%%%%%%%%%%%%%%%%% Results figure (2-column) %%%%%%%%%%%%%%%%%%%%%%%%%%%%%
\begin{figure*}[!t]
\centering
\includegraphics[width=0.85\linewidth]{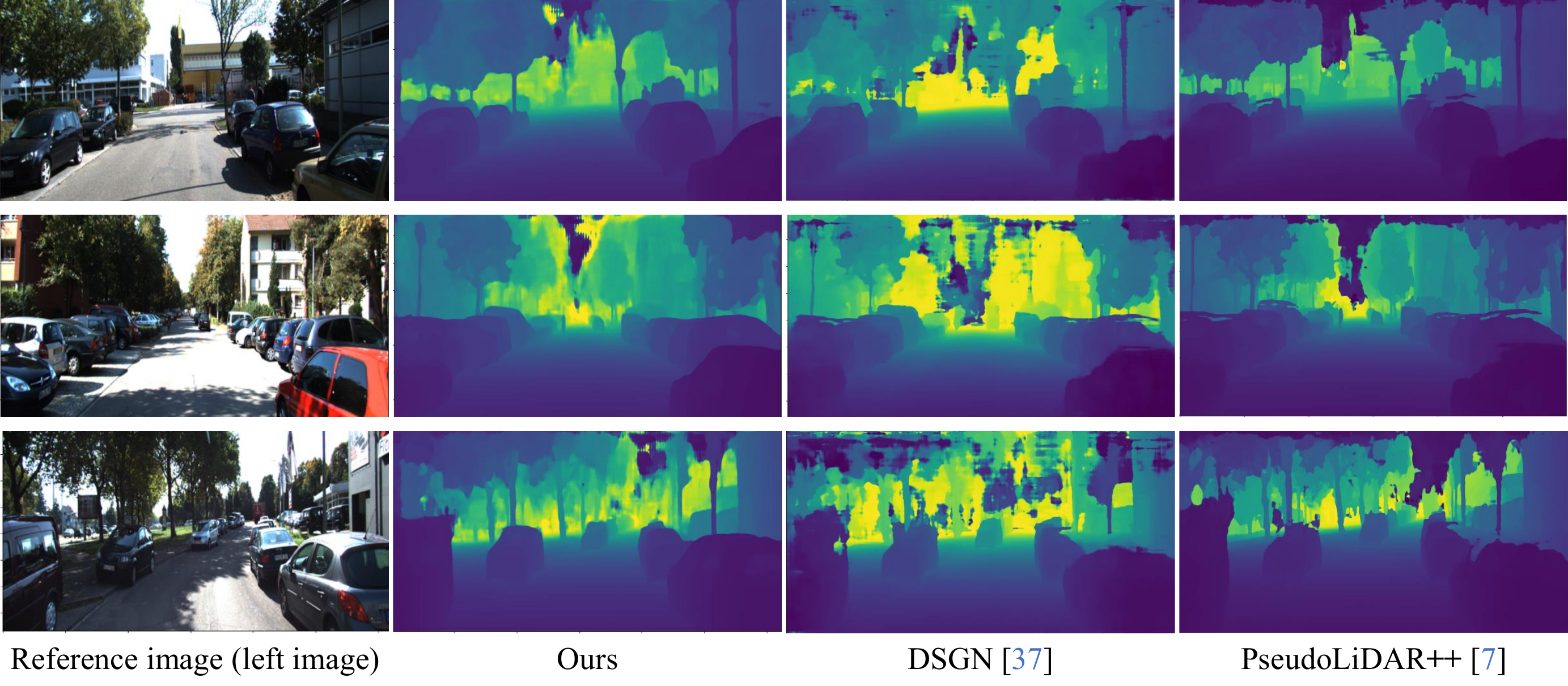}
\vspace{-3mm}
\caption{\textbf{Qualitative results of ours and the related works~\cite{DSGN,pseudo_lidar++} on the KITTI dataset.}
}
\label{fig:fig-depth}
\vspace{-1\baselineskip}
% \vspace{-2mm}
\end{figure*}
%%%%%%%%%%%%%%%%%%%% Results figure (1-column) %%%%%%%%%%%%%%%%%%%%%%%%%%%%%
{
% \begin{figure}[!t]
% \centering
% \includegraphics[width=1.0\linewidth]{fig_depth/fig_depth_00.pdf}
% % \vspace{-3mm}
% \caption{\textbf{Qualitative results of ours and the related works~\cite{DSGN,pseudo_lidar++} on the KITTI dataset.}}
% \label{fig:fig-depth}
% % \vspace{-1\baselineskip}
% \end{figure}
}

\subsection{Evaluation} \label{subsection:evaluation}
\noindent \textbf{Metric.} \ 
We quantitatively evaluate the proposed method for depth estimation (disparity) and 3D object detection.
For the depth estimation, we utilize the depth metric proposed by Eigen~\etal~\cite{eigen_mono_depth}.
For the disparity evaluation, we employ the official disparity metric from the KITTI stereo benchmark~\cite{kitti-stereo}.
As for the detection evaluation, we follow the rules provided by the KITTI object benchmark~\cite{kitti-object}, where we only consider the single label (vehicle object).
The detection evaluation is divided into three cases: easy, moderate, and hard.
We measure the Average Precision~(AP) of the 3D detection and the bird’s eye view 
(BEV) detection, where the threshold of the intersection of union~(IoU) is $0.7$.

We compare our algorithm with the recent deep learning-based depth estimation methods~\cite{monodepth,cspn,pseudo_lidar,psmnet} (both monocular-based and stereo-based methods) as shown in~\Tref{table:kitti-eigen-metric}.
In particular, our evaluation focuses on the quality of the depth rather than the disparity. 
Given the camera focal length and baseline, both depth and disparity mathematically indicate the same 3D measurement. 
However, while the depth map expresses the 3D measurement in the metric space, the disparity represented in the pixel domain has a non-linear mapping in the metric space. Therefore, the disparity metric may not properly reflect the error for distant objects.
In this regard, recent research works~\cite{pseudo_lidar,pseudo_lidar++} addressed the problem of the quality of disparity in view of the depth. 
Thus, to validate the practical usage for advanced tasks working on metric space, such as 3D object detection~\cite{pseudo_lidar,monodepthdetection}, we evaluate our method with respect to the quality of depth.

\vspace{1mm} \noindent \textbf{Comparison.} \
As~in~\Tref{table:kitti-eigen-metric} and~\Tref{table:virtual-kitti-metric}, we compare our method with the recent works for the depth estimation~\cite{eigen_mono_depth,dorn} and the stereo matching~\cite{pseudo_lidar,psmnet,DSGN}. 
These methods can be divided into monocular-based, mono-video, and stereo-based methods.
Since stereo matching methods train their network with the different benchmarks and different evaluations, we re-train the publicly accessible methods~\cite{pseudo_lidar,psmnet} to validate the performance with respect to the quality of the depth. 
Generally, the stereo-based disparity estimation methods show a higher depth accuracy than those of the monocular-based approaches; in particular, our method further improves the accuracy with a large margin. 
In addition, \Fref{fig:fig-depth} shows that our method has a lower depth error around object boundaries, even for distant objects. 

%%% DO NOT ERASE IT !!!!
%
% On the disparity metric, our approach shows relatively low accuracy on the KITTI stereo benchmark compared to the state-of-the-art methods that rely on similar baseline networks~\cite{gcnet,psmnet}, as shown in~\Tref{table:disparity-metric} and~\Tref{table:virtual-kitti-metric}.  
% However, the proposed method shows a reasonable result of D1-bg. This metric is computed by the disparity metric from the farther distant pixels (\ie,~background), where the objects (\eg,~vehicles) are usually located. Thus, the results are consistent with the intention of the stereo object matching network.

This result still supports the quality of our estimated depth. 
Thus, our detection-based 3D embedding is helpful to refine the cost volume, and it creates an increase in the depth quality, typically for faraway objects.
We also evaluate the performance of the 3D object detection with camera-based approaches~\cite{mono_rpn,stereo3d,triangulation_detection,pseudo_lidar,DSGN} in~\Tref{table:kitti-object-metric}. 
Despite the gap of 3D object detection accuracy between ours and related studies,
%
% our intention of usage of detection loss for stereo matching task has been proved \FR{"our hypothesis (including 3D object detection as an auxiliary task to boost depth estimation performance) is validated" instead of "our intention of usage of detection loss for stereo matching task has been proved".}. 
our hypothesis (including 3D object detection as an auxiliary task to boost depth estimation performance) is validated.
%
%Nonetheless, \FR{remove "Nonetheless"} 
%
% We provide further analysis of \FR{"regarding the" instead of "of"} detection performance \FR{"in the supplementary material" instead of ", which is included in the supplementary material"}, which is included in the supplementary material.
We provide further analysis regarding the detection performance in the supplementary material.

%------------------------------------------------------------------------
\subsection{Ablation Study}
\label{subsection:ablation-study}

% %% ablation : RoI sampling (w/o RMSE log) (simple naming) %%%%%%%%%%%%%%
\begin{table}[!t]
\centering
\caption{\textbf{Ablation study of detection modules and the RoI sampling in the KITTI dataset.} 
We intentionally omit the components of detection modules (\ie, 3D RPN and Header) and our RoISelect. 
We details our RoISelect into the two components, Deep sample and Select~(\cf,~Figs.~\ref{fig:sampling}(\textcolor{blue}{b, c})). 
Method 1 is the baseline~\cite{psmnet} of our network. 
It shows that the high-level perception task (\ie,~3D object detection) positively affects the low-level vision task (depth estimation).
}
\vspace{-2mm}
\resizebox{1.0\linewidth}{!}{
\begin{tabular}{|c|c|c|c|c|c|c|c|}
\hline
\multirow{4}{*}{Method} & \multicolumn{4}{c|}{Preserve (\checkmark)} & \multirow{2}{*}{Depth} & \multicolumn{2}{c|}{\multirow{2}{*}{Detection}} \\ \cline{2-5}

& \multicolumn{2}{c|}{Detection} & \multicolumn{2}{c|}{RoI Select} & & \multicolumn{2}{c|}{} \\ \cline{2-8}

& 3D & \multirow{2}{*}{Header} & Deep & \multirow{2}{*}{Select} & \multirow{2}{*}{RMSE} & $\text{AP}_\text{BEV}$ & $\text{AP}_\text{3D}$ \\ 

& RPN & & sample & & & (Moderate) & (Moderate)  \\ \hline

1 & & & & & 2.847 & - & - \\ \hline

2 & $\checkmark$ & & & & 1.938 & - & -  \\ \hline

3 & $\checkmark$ & $\checkmark$ & & & 1.934  & 21.77 & 13.72 \\ \hline

4  & $\checkmark$ & $\checkmark$ & $\checkmark$ & & 1.924 & 25.80 & 16.90 \\ \hline

5 & $\checkmark$ & $\checkmark$ & $\checkmark$ & $\checkmark$ & \textbf{1.842} & \textbf{30.92} & \textbf{19.88} \\ \hline

\end{tabular}
}
\vspace{-1\baselineskip}
\label{table:ablation-sampling}
\end{table}

\vspace{1mm} \noindent \textbf{RoISelect.} \
We set the baseline (Method~\textcolor{blue}{1}~in~\Tref{table:ablation-sampling}) as the pure disparity estimation network, PSMnet~\cite{psmnet},
which does not contain detection networks (\ie, the 3D-RPN and the header network).
In \Tref{table:ablation-sampling}, we evaluate the performance of each method in the KITTI-split~\cite{eigen_mono_depth} validation set, where our network (Method~\textcolor{blue}{5}) shows the best performance in both depth estimation and detection. From this, we can deduce that our sampling strategy (Select in~\Tref{table:ablation-sampling}) can handle occlusion cases by ignoring uncertain matching costs. 

\vspace{1mm} \noindent \textbf{Fusion-by-occupancy.} \
We first analyze the influence of the different types of volumes (\ie, cost volume $\mathcal{V}$ and cost aggregation $\mathcal{A}$) for the input of the detection modules.
As shown in~\Tref{table:ablation-fusion} (Methods~\textcolor{blue}{6} and \textcolor{blue}{7}), cost volume $\mathcal{V}$ is the appropriate voxel representation to include the detection-based 3D contextual information. 
We deduce that channel-wise information in cost volume $\mathcal{V}$ (\ie voxel features) is crucial for the detection task, as channel-wise information was crucial in image recognition~\cite{alexnet,vgg,resnet}.

We extend our investigation by fusing $\mathcal{V}$ with the estimated disparity map.
In Method~\textcolor{blue}{8}, we extract the 2D RoIs ($16\times16$) from the disparity map $\widetilde{\mathcal{D}}$ and repeatedly stack the 2D RoIs along the disparity axis ($16\times16\times16$).
Then, we apply the same process, as shown in~\Fref{fig:fusion}.
The last experiment (Method~\textcolor{blue}{9}) is the one that adopts our fusion-by-occupancy (denoted as 3D in~\Tref{table:ablation-fusion}).
The resulting scores show that our method outperforms the 2D fusion method in the evaluation of the two tasks.
These results demonstrate that the way of aligning the dimensions is an important issue for the cost volume-based detection approach.

%%%%%%%%%%%%%%%%%%%%% ablation : 2D-3D fusion (w/o RMSE log) (simple naming) %%%%%%%%%%%%%%%%%%%%
\begin{table}[!t]
\centering
\caption{\textbf{Ablation study of the 2D-3D fusion in different volumes.}
We validate the influence of type of fusion under two types of volume, cost volume $\mathcal{V}$ and cost aggregation $\mathcal{A}$. 
It shows that $\mathcal{V}$ is the appropriate volume for fusion. Moreover, our fusion-by-occupancy (denoted as 3D) increases accuracy in both depth and detection.
}
\vspace{-2mm}
\resizebox{1.0\linewidth}{!}{
\begin{tabular}{|c|c|c|c|c|c|c|c|}
\hline

\multirow{4}{*}{Method} & \multicolumn{4}{c|}{Preserve (\checkmark)} & \multirow{2}{*}{Depth} & \multicolumn{2}{c|}{\multirow{2}{*}{Detection}} \\ \cline{2-5}

& \multicolumn{2}{c|}{Input volume} & \multicolumn{2}{c|}{Fusion type} & & \multicolumn{2}{c|}{} \\ \cline{2-8}

& \multirow{2}{*}{costV} & \multirow{2}{*}{costA} & \multirow{2}{*}{2D} & 3D & \multirow{2}{*}{RMSE} & $\text{AP}_\text{BEV}$ & $\text{AP}_\text{3D}$ \\ 

& & & & (ours) & & (Moderate) & (Moderate)  \\ \hline

% & \multicolumn{4}{c|}{} & (Lower the better) & \multicolumn{2}{c|}{(Higher the better)} \\ \cline{2-8}

% & \multicolumn{2}{c|}{Type of volume} & \multicolumn{2}{c|}{Type of fusion} & \multirow{2}{*}{RMSE} & $\text{AP}_\text{BEV}$ & $\text{AP}_\text{3D}$ \\ \cline{2-5}

% & costV & costA & \text{ } 2D \text{ } & 3D (ours) & & (Moderate) & (Moderate)  \\ \hline

6 & \checkmark & & &  & 0.139  & 28.94 & 17.96 \\ \hline

7 & & \checkmark & &  & 0.116  & 25.73 & 12.08 \\ \hline

8 & \checkmark & & \checkmark &  & 0.113 & 24.49 & 15.71 \\ \hline

9 & \checkmark & & & \checkmark  & \textbf{0.111} & \textbf{30.92} & \textbf{19.88} \\ \hline

\end{tabular}
}
\vspace{-1\baselineskip}
\label{table:ablation-fusion}
\end{table}

% =======================================
% Conclusion
% =======================================
\section{Conclusion}

We have proposed a novel stereo object matching network.  
In particular, we investigated how to embed 3D object-level information in the stereo matching framework.
In contrast to existing techniques, we exploit the cost volume structure as a semi-3D space in which the 3D object detection task can be included as an auxiliary task.  
However, this integration is not straightforward and requires the use of specific architectures: RoISelect and fusion-by-occupancy to handle the specificity of the cost volume.
This strategy allows us to seamlessly embed 3D object-level contextual information in the cost volume. 
With this embedding, we can reduce the disparity ambiguity at the objects' edges and achieve state-of-the-art performance over the recent approaches.

% By this embedding, we can reduce the disparity ambiguity at the objects' edges and achieve state-of-the-art performance on the KITTI dataset and Virtual-KITTI dataset.  

% \section*{ACKNOWLEDGMENT}
% This work was supported by Institute of Information \& communications Technology Planning \& Evaluation (IITP) grant funded by the Korea government (MSIT)
% (No.2020-0-01336, Artificial Intelligence Graduate School Program (UNIST)).
% This research was also supported by the Shared Sensing for Cooperative Cars Project funded by Bosch (China) Investment Ltd.

\bibliographystyle{IEEEtran}
\bibliography{egbib}
 
\end{document}